\documentclass[11pt,a4paper]{article}
\usepackage[hyperref]{naaclhlt2019}
\usepackage{times}
\usepackage{latexsym}

\usepackage{bm}
\usepackage{url}
\usepackage{epsfig}
\usepackage{amsmath}
\usepackage{amsfonts,amssymb}
\usepackage{subcaption}
\usepackage{graphicx}
\usepackage{bm}
\usepackage{soul}
\usepackage{xcolor}
\usepackage[group-separator={,},group-minimum-digits={3}]{siunitx}
\usepackage{booktabs}
\usepackage{multicol}
\usepackage{multirow}
\usepackage{algorithm}
\usepackage{algorithmicx}
\usepackage{algpseudocode}

\newcommand*{\affaddr}[1]{#1} 
\newcommand*{\affmark}[1][*]{\textsuperscript{#1}}
\newcommand*{\email}[1]{\texttt{#1}}

\aclfinalcopy 

\title{Learning Unsupervised Word Mapping by Maximizing Mean Discrepancy}

\author{Pengcheng Yang\affmark[1,2], Fuli Luo\affmark[2], Shuangzhi Wu\affmark[3], Jingjing Xu\affmark[2], Dongdong Zhang\affmark[3], Xu Sun\affmark[1,2]\\
\affaddr{\affmark[1]Deep Learning Lab, Beijing Institute of Big Data Research, Peking University}\\
\affaddr{\affmark[2]MOE Key Lab of Computational Linguistics, School of EECS, Peking University}\\
\affaddr{\affmark[3]Microsoft Research Asia}\\
\email{\{yang\_pc, luofuli, jingjingxu, xusun\}@pku.edu.cn}\\
\email{\{v-shuawu, dozhang\}@microsoft.com}\\
}

\date{}

\begin{document}
\maketitle
\begin{abstract}
Cross-lingual word embeddings aim to capture common linguistic regularities of different languages, which benefit various downstream tasks ranging from machine translation to transfer learning. Recently, it has been shown that these embeddings can be effectively learned by aligning two disjoint monolingual vector spaces through a linear transformation (word mapping). In this work, we focus on learning such a word mapping without any supervision signal. Most previous work of this task adopts parametric metrics to measure distribution differences, which typically requires a sophisticated alternate optimization process, either in the form of \emph{minmax game} or intermediate \emph{density estimation}. This alternate optimization process is relatively hard and unstable. In order to avoid such sophisticated alternate optimization, we propose to learn unsupervised word mapping by directly maximizing the mean discrepancy between the distribution of transferred embedding and target embedding. Extensive experimental results show that our proposed model outperforms competitive baselines by a large margin. 
\end{abstract}

\section{Introduction}
It has been shown that word embeddings are capable of capturing meaningful representations of words~\cite{mikolov2013efficient, pennington2014glove, BojanowskiGJM17}. Recently, more and more efforts turn to cross-lingual word embeddings, which benefit various downstream tasks ranging from machine translation~\cite{lample2017unsupervised} to transfer learning~\cite{zhouWX16}.

Based on the observation that monolingual word embeddings share similar geometric properties across languages~\cite{mikolov2013exploiting}, the underlying idea is to align two disjoint monolingual vector spaces through a linear transformation. \citet{xing15} further empirically demonstrate that the results can be improved by constraining the desired linear transformation as an orthogonal matrix, which is also proved theoretically by~\citet{smith2017offline}. 

Recently, increasing effort of research has been motivated to learn word mapping without any supervision signal. One line of research focuses on designing heuristics~\cite{KondrakHN17} or taking advantage of the structural similarity of monolingual embeddings~\cite{aldarmakiMD18,artetxe2018robust,hoshen}. However, these methods often require a large number of random restarts or additional skills such as re-weighting~\cite{artetxe2018generalizing} to achieve satisfactory results. 

Another line strives to learn word mapping by matching distribution of transferred embedding and target embedding. For instance,~\citet{zhangadv} and ~\citet{conneau2017word} implement the word mapping as generator in GAN~\cite{goodfellow2014generative}, which is essentially a \emph{minmax game}. \citet{zhangearth} and~\citet{xu2018unsupervised} adopt the Earth Mover's distance~\cite{RubnerTG98} and Sinkhorn distance~\cite{Cuturi13} as the optimized metrics respectively, both of which require intermediate \emph{density estimation}. Although this line exhibits relatively excellent performance, both the \emph{minmax game} and the intermediate \emph{density estimation} require alternate optimization. However, such a sophisticated alternate optimization process tends to cause a hard and unstable optimization problem~\cite{grave2018unsupervised}.

In this paper, we follow the core idea of the second line basing on distribution-matching and combine it with the first line simultaneously. Different from the previous work requiring sophisticated alternate optimization, we propose to learn unsupervised word mapping by directly maximizing the mean discrepancy between the distribution of transferred embedding and target embedding. The \emph{Maximum Mean Discrepancy} (MMD) is a non-parametric metric, which measures the difference between two distributions. Compared with other parametric metrics, it does not require any intermediate \emph{density estimation}, leading to a more stable optimization problem. Besides, in order to alleviate the initialization sensitive issue of the distribution-matching, we take advantage of the structural similarity of monolingual embeddings~\cite{artetxe2018robust} to learn the initial word mapping to provide a warm-up start. 

The main contributions of this paper are concluded as follows:
\begin{itemize}
	\item We systematically analyze the drawbacks of the current models for the task of learning unsupervised word mapping.
    \item We propose to learn unsupervised word mapping by means of non-parametric maximum mean discrepancy, which avoids a relatively sophisticated alternate optimization process.
    \item Extensive experimental results show that our approach outperforms competitive baselines by a large margin on two benchmark tasks.
\end{itemize}

\section{Proposed Method}
\subsection{Overview}
Here we define some notations and describe the task of learning word mapping between different languages without any supervision signal. Let ${\mathcal X} = \{x_i\}_{i=1}^n$ and ${\mathcal Y} = \{y_i\}_{i=1}^m$ be two sets of $n$ and $m$ pre-trained monolingual word embeddings with dimensionality $d$, which come from the source and target language, respectively. Our goal is to learn a word mapping $\mathbf{W} \in \mathcal{O}_d$ so that for any source word embedding $x$, $\mathbf{W}x$ lies close to the embedding of its target language translation $y$. Here $\mathcal{O}_d$ is the space composed of all $d \times d$ orthogonal matrices. From the perspective of distribution-matching, the task of learning word mapping can be modeled as finding an optimal orthogonal matrix $\mathbf{W}^\star$ to make the distributions of $\mathbf{W}\mathcal{X}$ and $\mathcal{Y}$ as close as possible.

\subsection{MMD-Matching}
Concisely, \emph{Maximum Mean Discrepancy} (MMD) measures the difference between two distributions based on the \emph{Reproducing Kernel Hilbert Space} (RKHS) $\mathcal{H}$. Let $\mathcal{P}$ and $\mathcal{Q}$ represent the distribution of $\mathbf{W}\mathcal{X}$ and $\mathcal{Y}$, respectively, i.e., $\mathbf{W}x \sim \mathcal{P}$ and $y \sim \mathcal{Q}$. Then, the difference between the distributions $\mathcal{P}$ and $\mathcal{Q}$ can be characterized by:
\begin{equation}
\label{equ1}
{\rm MMD}(\mathcal{P}, \mathcal{Q}) = \Vert \mathbb{E}_{\mathbf{W}x \sim \mathcal{P}}[\phi(\mathbf{W}x)] - \mathbb{E}_{y \sim \mathcal{Q}} [\phi(y)] \Vert_\mathcal{H} 
\end{equation}
where $\phi(\cdot)$ is the feature mapping. ${\rm MMD}(\mathcal{P}, \mathcal{Q})$ reaches its minimum only when the distributions $\mathcal{P}$ and $\mathcal{Q}$ match exactly. Therefore, in order to match the distribution of transferred embedding and target embedding as exactly as possible, the underlying linear mapping $\mathbf{W}$ can be learned by solving the following optimization problem:
\begin{equation}
\label{equ2}
\min_{\mathbf{W} \in \mathcal{O}_d} {\rm MMD}(\mathcal{P}, \mathcal{Q})
\end{equation}
By means of \emph{kernel trick}~\cite{gretton2012kernel}, the MMD of the distributions $\mathcal{P}$ and $\mathcal{Q}$ can be calculated as:
\begin{align}
{\rm MMD}^2(\mathcal{P}, \mathcal{Q}) = & \ \mathbb{E}_{\mathbf{W}x \sim \mathcal{P},\mathbf{W}x' \sim \mathcal{P}} [k(\mathbf{W}x, \mathbf{W}x')] \nonumber \\
+ & \ \mathbb{E}_{y \sim \mathcal{Q},y' \sim \mathcal{Q}} [k(y,y')] \label{equ3} \\
- & 2\mathbb{E}_{\mathbf{W}x \sim \mathcal{P}, y \sim \mathcal{Q}} [k(\mathbf{W}x, y)]  \nonumber
\end{align}
where $k(\cdot,\cdot): \mathbb{R}^d \times \mathbb{R}^d \mapsto \mathbb{R}$ is the kernel function, such as polynomial kernel or Gaussian kernel. At the training stage, Eq.(\ref{equ3}) can be estimated by the sampling method, which is formulated as:
\begin{align}
{\rm MMD}^2(\mathcal{P}, \mathcal{Q}) = & \frac{1}{B^2}\Big[\sum_{i,j=1}^Bk(\mathbf{W}x_i, \mathbf{W}x_j) \\
- & 2\sum_{i,j=1}^Bk(\mathbf{W}x_i, y_j) + \sum_{i,j=1}^Bk(y_i, y_j)\Big] \nonumber 
\end{align}
where $B$ is the size of mini-batch. 

In order to maintain the orthogonality of $\mathbf{W}$ during training~\cite{smith2017offline}, we adopt the same update strategy proposed in~\citet{cisse2017parseval}. In detail, we replace the original update of the matrix $\mathbf{W}$ with the following update rule: 
\begin{equation}
\label{equ4}
\mathbf{W} := (1 + \beta) \mathbf{W} - \beta (\mathbf{W} \mathbf{W}^T)\mathbf{W}
\end{equation}
where $\beta$ is a hyper-parameter. After the optimization process of matching the distributions $\mathcal{P}$ and $\mathcal{Q}$ converges, we use the iterative refinement~\cite{conneau2017word,artetxe2018robust} to further improve results.

\begin{table*}[t]
\centering
\small
\setlength{\tabcolsep}{9.5pt}
\begin{tabular}{l|c c c c c c c c}
\toprule
 Methods & FR-EN & EN-FR & DE-EN & EN-DE & ES-EN & EN-ES & IT-EN & EN-IT \\ 
 \midrule
 \textbf{Supervised:} & & & & & & & & \\
\citet{mikolov2013efficient} & 71.33 & \textbf{82.20} & 61.93 & \textbf{73.07} & 74.00 & \textbf{80.73} & 68.93 & \textbf{77.60} \\
 \citet{xing15} & 76.33 & 78.67 & 67.73 & 69.53 & 77.20 & 78.60 & 72.00 & 73.33 \\
 \citet{ShigetoSHSM15} & \textbf{79.93} & 73.13 & \textbf{71.07} & 63.73 & \textbf{81.07} & 74.53 & \textbf{76.47} & 68.13 \\
  \citet{ZhangGBJ16} & 76.07 & 78.20 & 67.67 & 69.87 & 77.27 & 78.53 & 72.40 & 73.40 \\
 \citet{ArtetxeLA16} & 77.73 & 79.20 & 69.13 & 72.13 & 78.27 & 80.07 & 73.60 & 74.47 \\
 \citet{artetxeLA17} & 74.47 & 77.67 & 68.07 & 69.20 & 75.60 & 78.20 & 70.53 & 71.67 \\ 
 \midrule
 \textbf{Unsupervised:} & & & & & & & & \\
\citet{zhangadv} & * & 57.60 & 40.13 & 41.27 & 58.80 & 60.93 & 43.60 & 44.53\\
 \citet{conneau2017word} & 77.87 & 78.13 & 69.73 & 71.33 & 79.07 & 78.80 & 74.47 & 75.33 \\
 \citet{xu2018unsupervised} & 75.47 & 77.93 & 67.00 & 69.33 & 77.80 & 79.53 & 72.60 & 73.47 \\ 
 Ours & \textbf{78.87} & \textbf{78.40} & \textbf{70.33} & \textbf{71.53} & \textbf{79.33} & \textbf{79.93} & \textbf{74.73} & \textbf{75.53} \\
 \bottomrule
\end{tabular}
\caption{The accuracy of different methods in various language pairs on the bilingual lexicon induction task. The best score for each language pair is bold for the supervised and unsupervised categories, respectively. For the baseline~\citet{zhangadv}, we adopt the most commonly used unidirectional transformation model. ``*'' means that the model fails to converge and hence the result is omitted.}
\label{tab1}
\end{table*}

\subsection{Compression and Initialization}
At the training stage, Eq.(\ref{equ3}) is estimated by the sampling method. The bias of estimation directly determines the accuracy of calculation of the MMD. A reliable estimation of Eq.(\ref{equ3}) generally requires the size of the mini-batch to be proportional to the dimension. Therefore, we adopt a \emph{compressing network} implemented as a multilayer perceptron to map all embeddings into a lower feature space. Experimental results show that the use of \emph{compressing network} can not only improve the performance of the model, but also provide significant computational savings.

Besides, we find that the training of the model is sensitive to the initialization of word mapping. An inappropriate initialization tends to cause the model to stuck in poor local optimum. The same sensitivity issue is also observed by~\citet{zhangearth,aldarmakiMD18,xu2018unsupervised}. Therefore, we take advantage of the structural similarity of embeddings to provide the initial setting for our MMD-matching process. Readers can refer to~\citet{artetxe2018robust} for detailed approach.

\section{Experiments}

\subsection{Evaluation Tasks}
We evaluate our proposed model on two representative benchmark tasks: bilingual lexicon induction and cross-lingual word similarity prediction. 

\paragraph{Bilingual lexicon induction} 
The goal of this task is to retrieve the translation of given source word. We use the lexicon constructed by~\citet{conneau2017word}. Here we report accuracy with \emph{nearest neighbor retrieval} based on cosine similarity.

\paragraph{Word similarity prediction}
This task aims to measure how well the predicted cross-language word cosine similarities correlate with the human-labeled scores. Following~\citet{conneau2017word}, we use the SemEval 2017 competition data. We report Pearson correlation between the predicted similarity scores and the human-labeled scores over testing word pairs for each language pair. 

\subsection{Experiment Settings}
For both evaluation tasks, we use publicly available 300-dimensional \emph{fastText} word embeddings trained on Wikipedia. The compressing network is used to map the original 300-dimensional embeddings to 50-dimensional. The batch size is set to 1280 and $\beta$ in Eq.(\ref{equ4}) is set to 0.01. We use a  mixture of 10 isotropic Gaussian (RBF) kernels with different bandwidths $\sigma$ as in \citet{LiSZ15}. The Adam optimizer is used to minimize the final objective function. The learning rate is initialized to $0.0003$ and it is halved after every training epoch. We adopt the unsupervised criterion proposed in~\citet{conneau2017word} both as a stopping criterion and to select the best hyper-parameters.

\subsection{Results}
The experimental results of our approach and the baselines on the bilingual lexicon induction task are shown in Table~\ref{tab1}. Results show that our proposed model outperforms all unsupervised baselines by a large margin,  which shows that the use of MMD is of great help to improve the quality of word mapping. Compared with the supervised methods, it is gratifying that our approach also achieves completely comparable performance. 

Table~\ref{tab2} summarizes the performance of all methods on the cross-lingual word similarity prediction task. Similar to results in Table~\ref{tab1}, our proposed model still achieves the best performance compared with the unsupervised baselines, which is also highly comparable to competitive supervised methods.

\begin{table}[t]
\centering
\footnotesize
\setlength{\tabcolsep}{3.7pt}
\begin{tabular}{l|c c c c}
\toprule
 Methods & EN-ES & EN-FA & EN-DE & EN-IT \\ 
 \midrule
 \textbf{Supervised:} & & & & \\
\citet{mikolov2013efficient} & 0.71 & 0.68 & 0.71 & 0.71 \\
 \citet{xing15} & 0.71 & 0.69 & 0.72 & 0.72 \\
 \citet{ShigetoSHSM15} & \textbf{0.72} & 0.69 & 0.72 & 0.71 \\
 \citet{ZhangGBJ16} & 0.71 & 0.69 & 0.71 & 0.71 \\
 \citet{ArtetxeLA16} & \textbf{0.72} & \textbf{0.70} & \textbf{0.73} & \textbf{0.73} \\
 \citet{artetxeLA17} & 0.70 & 0.67 & 0.70 & 0.71 \\
 \midrule
  \textbf{Unsupervised:} & & & & \\
\citet{zhangadv} & 0.68 & 0.65 & 0.69 & 0.69 \\
 \citet{conneau2017word} & 0.71 & \textbf{0.68} & \textbf{0.72} & 0.71 \\
 \citet{xu2018unsupervised} & 0.71 & 0.67 & 0.71 & 0.71 \\ 
 Ours & \textbf{0.72} & 0.67 & \textbf{0.72} & \textbf{0.72} \\
 \bottomrule
\end{tabular}
\caption{Comparison between our approach and all baselines on the word similarity prediction task. Pearson correlation between the predicted similarity scores and the human-labeled scores is reported.}
\label{tab2}
\end{table}

\section{Analysis and Discussions}
Here we perform further analysis on the model and experiment results.

\subsection{Ablation Study}
Here we perform an ablation study to understand the importance of different components of our approach. Table~\ref{tab3} presents the best performance obtained by multiple versions of our model with some missing components: the MMD-matching, the refinement, and the initialization. 

The most critical component is initialization, without which our model will fail to converge. This initialization sensitivity issue is ingrained and difficult to eliminate in the optimization of some metrics~\cite{aldarmakiMD18,xu2018unsupervised}. Besides, as is shown in Table~\ref{tab3}, the final refinement can bring a significant improvement in performance. What we need to emphasize is that although the missing of MMD-matching brings the weakest decline in performance, it is still a key component to guide the model to learn a better final word mapping.

\begin{table}[t]
\centering
\footnotesize
\setlength{\tabcolsep}{4pt}
\begin{tabular}{l|c c c c}
\toprule
 Models & EN-ES & EN-FR & EN-DE & EN-IT \\ 
 \midrule
\emph{Full model} & 79.93 & 78.40 & 71.53 & 75.53 \\
\midrule
\emph{w/o MMD-matching} & 71.60 & 72.53 & 68.20 & 71.40 \\
\emph{w/o Refinement} & 55.80 & 65.27 & 61.00 & 58.67 \\
\emph{w/o Initialization} & * & * & * & * \\
 \bottomrule
\end{tabular}
\caption{Ablation study on the bilingual lexicon induction task. ``*'' means that the model fails to converge and hence the result is omitted.}
\label{tab3}
\end{table}

\subsection{Error Analysis}
\begin{figure}[tb]
\centering
\includegraphics[width=0.95\linewidth]{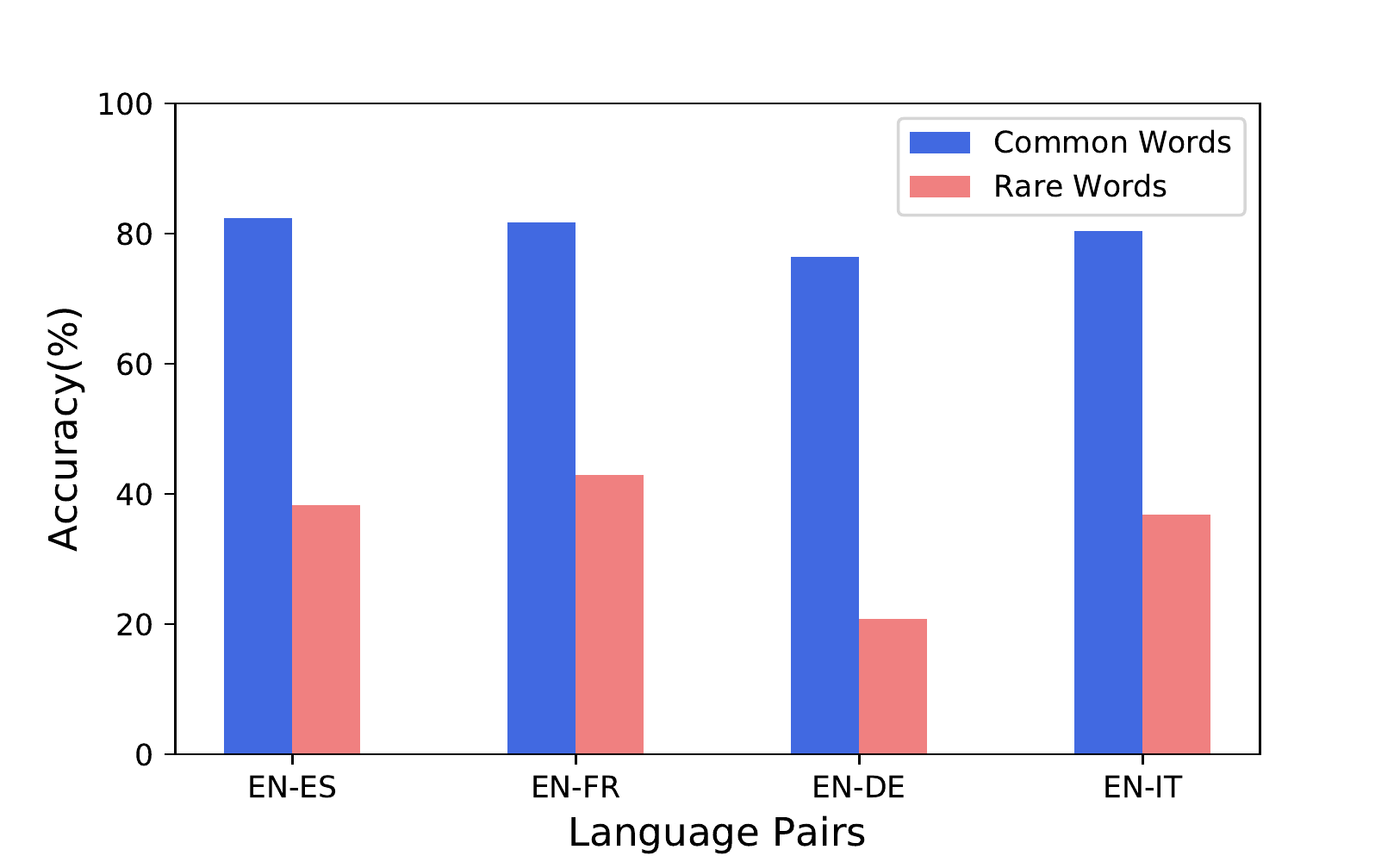}
\caption{The performance of our approach in common words and rare words on the bilingual lexicon induction task. Common words are the most frequent 20,000 words, and the remaining are regarded as rare words.}
\label{error_analysis}
\end{figure}

In the experiment, we find that all methods exhibit relatively poor performance when translating rare words on the bilingual lexicon induction task. Figure~\ref{error_analysis} shows the performance of our approach on the common word pairs and rare word pairs, from which we can see that the performance is far worse when the model translates rare words.

Since the pre-trained monolingual embeddings provide the cornerstone for learning unsupervised word mapping, the quality of monolingual embeddings directly determines the quality of word mapping. Due to the low frequency of rare words, the quality of their embeddings is lower than that of common words. This makes the isometric assumption~\cite{artetxe2018robust} more difficult to satisfy on rare words, leading to poor performance of all methods on rare word pairs. Improving the quality of cross-lingual embeddings of rare words is expected to be explored in future work.

\section{Conclusion}
In this paper, we propose to learn unsupervised word mapping between different languages by directly maximizing the mean discrepancy between the distribution of transferred embedding and target embedding. The proposed model adopts non-parametric metric that does not require any intermediate density estimation, which avoids a relatively sophisticated and unstable alternate optimization process. Extensive experimental results show that the proposed model outperforms the baselines by a substantial margin.

\bibliography{naaclhlt2019}
\bibliographystyle{acl_natbib}

\end{document}